\documentclass[conference]{IEEEtran}
\IEEEoverridecommandlockouts
\usepackage{cite}
\usepackage{amsmath,amssymb,amsfonts}
\usepackage{algorithmic}
\usepackage[dvipdfmx]{gr aphicx}
\usepackage{graphicx}
\usepackage{textcomp}
\usepackage{xcolor}
\usepackage{algorithm} 
\usepackage{subfig}
\usepackage{ntheorem}
\usepackage{hyperref}
\usepackage{textcomp}
\usepackage{comment}
\usepackage{color}
\usepackage{soul}
\usepackage{url}
\usepackage[misc]{ifsym}

\theoremseparator{:}

\newboolean{showcomments}
\setboolean{showcomments}{true}
\ifthenelse{\boolean{showcomments}}
{\newcommand{\nb}[2]{
		\fcolorbox{black}{yellow}{\bfseries\sffamily\scriptsize#1}
		{\sf\small$\blacktriangleright$\textit{#2}$\blacktriangleleft$}
	}
	
}
{\newcommand{\nb}[2]{}
	
}

\makeatletter
\newcommand{\figcaption}[1]{\def\@captype{figure}\caption{#1}}
\newcommand{\tblcaption}[1]{\def\@captype{table}\caption{#1}}
\makeatother

\def\BibTeX{{\rm B\kern-.05em{\sc i\kern-.025em b}\kern-.08em
    T\kern-.1667em\lower.7ex\hbox{E}\kern-.125emX}}
\begin{document}
\bstctlcite{IEEEexample:BSTcontrol}

\newcommand\jialong[1]{\nb{Jialong}{#1}}
\newcommand\sherry[1]{\nb{Sherry}{#1}}
\newcommand\kenji[1]{\nb{Kenji}{#1}}

\title{Automatic Adaptation Rule Optimization\\via Large Language Models}

\author{
    \IEEEauthorblockN{
        Yusei Ishimizu\textsuperscript{1}, 
        Jialong Li\textsuperscript{2*},
        Jinglue Xu\textsuperscript{3},
        Jinyu Cai\textsuperscript{2}, 
        Hitoshi Iba\textsuperscript{3}, 
        Kenji Tei\textsuperscript{1}
    }
    \IEEEauthorblockA{\textit{\textsuperscript{1} School of Computing, Tokyo Institute of Technology, Tokyo, Japan}}
    \IEEEauthorblockA{\textit{\textsuperscript{2} Department of Computer Science and Engineering, Waseda University, Tokyo, Japan}}
    \IEEEauthorblockA{\textit{\textsuperscript{3} Graduate School of Information Science and Technology, University of Tokyo, Tokyo, Japan}}
        \IEEEauthorblockA{ishimizu.y.aa@m.titech.ac.jp, lijialong@fuji.waseda.jp, tei@c.titech.ac.jp}
    
}

\maketitle

\begin{abstract}
Rule-based adaptation is a foundational approach to self-adaptation, characterized by its human readability and rapid response. However, building high-performance and robust adaptation rules is often a challenge because it essentially involves searching the optimal design in a complex (variables) space. In response, this paper attempt to employ large language models (LLMs) as a optimizer to construct and optimize adaptation rules, leveraging the common sense and reasoning capabilities inherent in LLMs. Preliminary experiments conducted in SWIM have validated the effectiveness and limitation of our method.
\end{abstract}


\section{Introduction}
Self-adaptive systems are engineered to achieve system goals under dynamic conditions and changing environments. 
Rule-based adaptation, which is a fundamental approach where the adaptation logic is predefined through rules, offers two significant advantages. 
Firstly, the human-readable rules enhances the explainability of adaptation. 
Secondly, since reasoning and planning are not required at runtime, rule-based adaptation is suitable for scenarios that require quick responses.

However, designing high-performance and robust adaptation rules is often challenging. Essentially, it is an optimization problem within a complex design space. The development of adaptation rules typically involves two crucial steps. The initial step is the selection of variables, which includes both observed input variables and control variables, essentially designing the dimensions of the search space.
The subsequent step focuses on the searching and optimization of rules, essentially finding the optimal solution within the given search space, for instance, \cite{ztq_ICAC} employs reinforcement learning.

Recently, large language models (LLMs) have emerged as significant tools across various research fields. \cite{hollmann2023large} utilizes LLMs for automatic feature engineering, identifying semantically relevant variables from extensive datasets. Meanwhile, \cite{yang2024large} explores the potential of LLM-based optimization in traveling salesman problem.
Given the capabilities demonstrated in these studies, we believe LLMs have strong potential for automatically constructing and optimizing adaptive rules.

In this paper, we preliminarily explore the application of LLMs to develop and optimize adaptation rules. Drawing from the established MAPE-K reference architecture, we design a series of prompts tailored for the continuous iteration and optimization of these rules. The effectiveness of this method is initially validated in SWIM (Simulator for Web Infrastructure and Management) \cite{8595390}.

\section{Proposal}
\begin{figure}[thb!]
    \centering
    \includegraphics[width=1\linewidth]{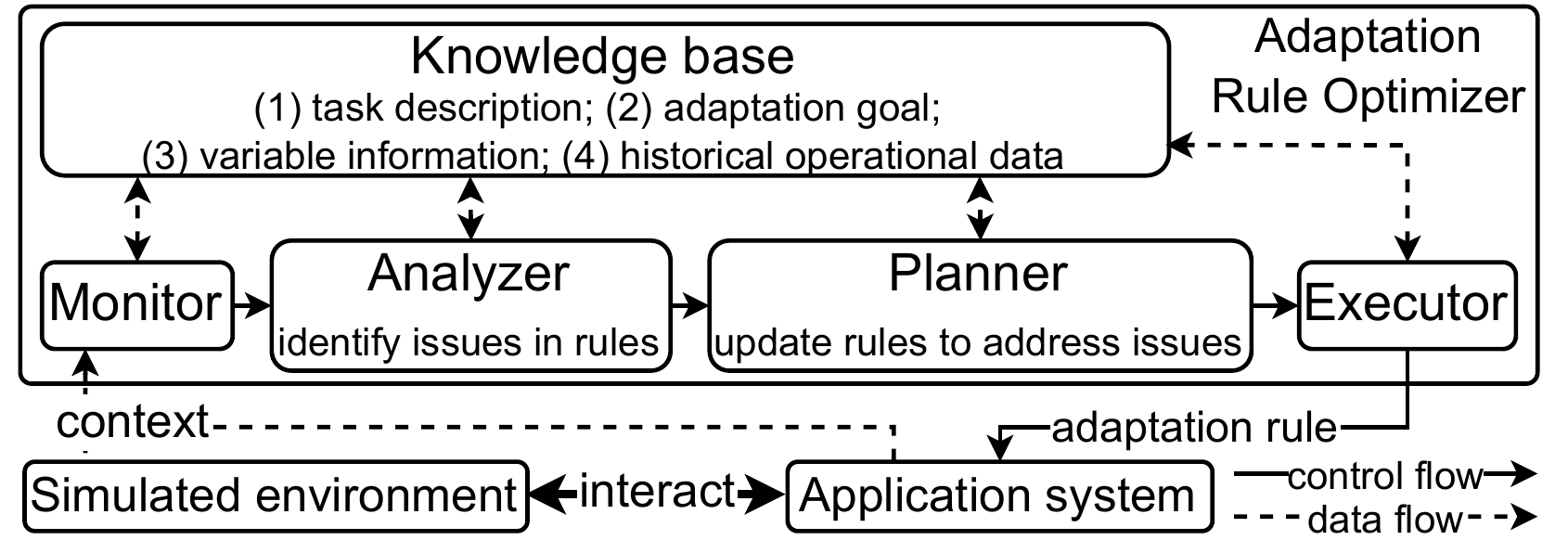}
    \caption{System Overview.} 
    \label{overview}
\end{figure}

\textbf{Overview}.
We consider the optimization of adaptation rules as processes of automatic computing\footnote{This paper only considers design-time rule construction; run-time rule evolution could be represented as the same process.}, and use the MAPE-K loop to outline our method, as shown in Fig.~\ref{overview}.
In this process, the monitor captures system and environment contexts from the interactions between the application system and its (simulated) environment, and incorporates them into the knowledge base.
Following this, the analyzer and the planner generate the updated adaptation rule, and the executor applies the new rules to the application system.
In the subsequent part of this section, we will focus primarily on the knowledge, analyzer, and planner components, as our method predominantly involves these three elements.

\textbf{Knowledge base}. When utilizing LLMs, it is essential to provide comprehensive context to support their reasoning. In knowledge base, we prepare four types of information crucial for rule construction. First, we include an introduction to the application domain, e.g., the operational logic of the application. Second, we define the adaptation goals, clarifying the objectives that the system aims to achieve. Third, we also describe the variables within the application, detailing both observable variables and those the system can control, i.e., defining the system’s observable and controllable spaces. Note that in rule construction, LLMs may only use some of these variables or even define new variables using existing ones. Lastly, we incorporate historical operational data, which could serve as feedback to enable LLMs to refine and iterate on adaptation rules.

\textbf{LLM-driven analyzer and planner}. 
The analyzer’s role is to identify potential issues within existing adaptation rules. For example, it might analyze the specific reasons for poor performance during certain timesteps and suggest which part of the rules to modify. Utilizing the insights generated by the analyzer, the planner then determines the necessary updates to the adaptation rules.  This could include introducing a new observable variable or suggesting adjustments to the values of control variables. It is important to note that both the analyzer and planner rely on the four types of information stored in the knowledge base, especially the historical operational data, to make well-informed decisions.

\section{Preliminary Evaluation}
\subsection{Experiment Setting}
The evaluation addresses the following research questions: What is the performance level of the adaptation rules designed and optimized by LLMs? How does the performance trend change with iterations of the adaptation rules?

\textbf{Target scenario}.
We utilize SWIM as the platform to simulate multi-tier web applications, where a load balancer supports multiple servers. Key variables include request arrival rate, server number (adjustable on demand) and the optional content dimmer.
The adaptation goal is to maximize a utility function primarily composed of revenue utility minus cost utility. 
We utilize Clarknet-105m-I70 scenario, which is a more challenging setting due to significant fluctuations in request arrivals per unit time.
We omit detailed explaination due to space constraints; interested readers can refer to \cite{8595390}.

\textbf{LLM and prompt settings}.
For the experiment, we employ advanced GPT-4 and excellent performance-cost DeepSeek-Coder-V2 (hereafter DS-Coder). 
Due to the uncertainty in LLMs' outputs, we conduct ten experiments for each LLM, and each experiment include ten iterations.
Furthermore, the adaptation rules are asked to be outputed in C++ language code format, allowing for direct execution within SWIM.

\subsection{Experiment Results and Discussion}

\begin{figure}[thb!]
    \centering
    \includegraphics[width=1\linewidth]{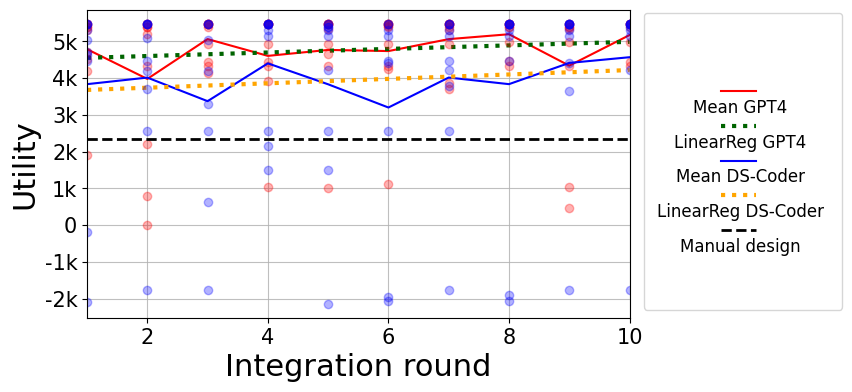}
    \caption{Results from ten experiments.} 
    \label{results}
\end{figure}

\textbf{Experiment Results}. Fig.~\ref{results} displays the results of the experiment, revealing several key points. First, both GPT-4 and DS-Coder shows superior performance (5.5k in the best cases) compared to the default manual-designed rules within SWIM. The deafult rules increase servers and adjust the dimmer when the average response time exceeds a threshold (vice versa), and use a total of 8 different variables in the conditionals. 
Second, due to the strong knowledge and reasoning capabilities of LLMs, the rules generated by LLMs demonstrated high performance even in the first round. In contrast, traditional search or optimization algorithms typically require multiple interations to achieve acceptable performance. Third, the results of linear regression indicate that the quality of adaptation rules improves with iterations, illustrating the feasibility of LLM-based optimization. However, performance does not necessarily improve with each iteration but shows significant fluctuations (even negative utilities), highlighting the trial-and-error process during iterations.

\textbf{Discussion and Limitations}. 
We believe our experimental results preliminarily demonstrate the effectiveness of using LLMs as optimizers for designing adaptation rules. However, the results also reveal the limitations of current approaches, namely the inefficiency of using LLMs directly as optimizers, as each iteration only explores one node in the design space. Even with LLMs possessing strong prior knowledge to guide the search, the search remains difficult in a multi-dimensional design space, as the slow response and high invocation cost of LLMs limit the number of iterations. This is also the reason why we do not choose existing optimization algorithms, such as evolutionary computation, as a baseline for comparison, because although existing algorithms often require more iterations for acceptable performance, their per-iteration cost is much lower. Therefore, a significant future research direction is to combine LLM's prior knowledge (as meta-heruistic) with existing search methods to achieve more efficient and economical optimization \cite{Jinyu_GECCO24}.

\section{Conclusion and Future Work}
\label{sec: conclusion}
In this paper, we utilize LLMs to automatically design and optimize adaptation rules. Preliminary experiments conducted in the SWIM environment have shown the effectiveness of our method. Future research will primarily focus on two areas. First, we aim to integrate existing optimization algorithms with LLMs to enhance the efficiency. Second, we plan to extend our method to the runtime phase, to allow the automatic evolution of adaptation rules under unforeseen conditions.

\bibliographystyle{IEEEtran}
\bibliography{bib}
\end{document}